\documentclass{article}

\usepackage{microtype}
\usepackage{graphicx}
\usepackage{booktabs}
\usepackage{hyperref}

\usepackage[accepted]{icml2026}

\makeatletter
\renewcommand{\ICML@appearing}{\textit{Accepted at ICML 2026 Workshop on Machine Learning for Audio, Seoul, South Korea. Copyright 2026 by the author(s).}}
\makeatother

\usepackage{amsmath,amssymb,amsfonts}
\usepackage{multirow}
\usepackage{xcolor}
\usepackage{makecell}
\usepackage{pgfplots}
\pgfplotsset{compat=1.17}

\icmltitlerunning{BoN TTS Family Alignment}

\begin{document}

\twocolumn[
  \icmltitle{Best-of-$N$ TTS Evaluation is Confounded by ASR Family Alignment}

  \icmlsetsymbol{equal}{*}

  \begin{icmlauthorlist}
    \icmlauthor{Taehyung Yu}{kaist}
    \icmlauthor{Seongjae Kang}{kaist}
  \end{icmlauthorlist}

  \icmlaffiliation{kaist}{KAIST, Daejeon, South Korea}

  \icmlcorrespondingauthor{Taehyung Yu}{taehyung.yu@kaist.ac.kr}

  \icmlkeywords{text-to-speech, best-of-N, verifier, evaluation, flow matching}

  \vskip 0.3in
]

\printAffiliationsAndNotice{}

\makeatletter
\gdef\@icmltitlerunning{Best-of-$N$ TTS Evaluation is Confounded by ASR Family Alignment}
\makeatother

\begin{abstract}
Best-of-$N$ (BoN) inference improves content consistency in zero-shot text-to-speech by selecting among multiple candidates with an automatic speech recognition (ASR) verifier. We identify an evaluation confound: the apparent quality of a verifier depends strongly on the ASR family used for evaluation. On LibriSpeech-PC with F5-TTS, verifier rankings vary substantially across Whisper, wav2vec~2.0, and HuBERT evaluators, while same-family verifier and evaluator pairs recover considerably more oracle headroom than cross-family pairs despite highly similar representations. This pattern suggests identity- or lineage-level coupling rather than general representational similarity. To mitigate this bias, we propose two \textbf{cross-family rank ensembles}, rank averaging and conjunctive max-rank. Both improve mean word error rate across independent evaluators without degrading automatic similarity or quality metrics, and the best ensemble achieves a $12\%$ relative WER reduction over F5-TTS at $N{=}10$. These findings motivate cross-evaluator triangulation as a more reliable default for reporting BoN TTS performance.
\end{abstract}

\section{Introduction}
\label{sec:intro}

Recent flow-matching zero-shot TTS systems---F5-TTS~\citep{f5tts}, E2 TTS~\citep{e2tts}, CosyVoice~2~\citep{cosyvoice2}, MaskGCT~\citep{maskgct}, Seed-TTS~\citep{seedtts}, NaturalSpeech~3~\citep{naturalspeech3}---produce speech that is indistinguishable from human recordings in naturalness and speaker similarity, yet still produce word-level content errors on a non-trivial fraction of utterances.
Best-of-$N$ (BoN) inference is the standard inference-time remedy: synthesize $N$ candidates with different random initializations and select the one whose ASR-decoded transcript matches the target text most closely.

BoN is reported to reduce WER by $10$--$30\%$ relative across recent flow-matching systems, but the literature is inconsistent on which \emph{verifier} ASR to use---reported choices include wav2vec~2.0~\citep{wav2vec2}, Whisper~\citep{whisper}, and its distillations~\citep{distilwhisper}---and the evaluation ASR is usually a single fixed model.
The choice of evaluator is a design decision that has so far received no systematic attention.

In this paper we run a four-way evaluator ablation spanning the Whisper~\citep{whisper}, wav2vec~2.0~\citep{wav2vec2}, and HuBERT~\citep{hubert} families (\S\ref{sec:setup}) over a shared set of BoN candidates. Our contributions are the following.
\begin{itemize}
\setlength{\itemsep}{0pt}\setlength{\parskip}{0pt}\setlength{\topsep}{0pt}
\item We document that BoN verifier rankings are systematically evaluator-dependent: the same generated outputs are ranked in opposing directions by evaluators from different ASR families, so the verifier preferred under one family can lose under another (\S\ref{sec:main}).
\item We quantify the magnitude of this confound on LibriSpeech-PC test-clean and show that the (verifier, evaluator) family pairing is a much larger lever on reported WER than the choice between common verifier checkpoints, while a large oracle headroom remains unexploited (\S\ref{sec:main}).
\item We rule out audio-encoder representation similarity (linear CKA) as the dominant explanation; the pattern is instead consistent with identity- or lineage-level coupling between verifier and evaluator, a speech analog of LLM-as-a-judge self-bias (\S\ref{sec:analysis}).
\item We propose \emph{cross-family rank ensembles} that select candidates by aggregating verifiers across ASR families, and recommend \emph{cross-evaluator triangulation}---reporting WER under at least two ASR families with disjoint training lineages---as a default reporting practice (\S\ref{sec:scaling}).
\end{itemize}

\section{Related Work}
\label{sec:related}

Inference-time remedies for flow-matching TTS---classifier-free guidance (CFG)~\citep{cfg}, BoN reranking~\citep{cosyvoice2,palle}, reverse inference~\citep{rio}---report no cross-evaluator validation; recent strong systems~\citep{seedtts,naturalspeech3} similarly tune verifiers against a single fixed ASR.
Speech preference optimization~\citep{speechalign} shares the same single-judge risk.
BoN scaling theory in LMs~\citep{scalingflaws,weaver} motivates verifier ensembling but has no TTS analog.
The structural precedent is LLM-as-judge self-bias~\citep{llmasjudge}; we document the speech analog.
Evaluator-driven shifts in spoken-language conclusions are established for ASR fairness~\citep{koenecke2020,feng2024} and for MOS-predictor cross-domain generalization~\citep{cooper2022,mittag2021nisqa,utmosv2}.

\section{Setup}
\label{sec:setup}

\textbf{Synthesis.}
We use the publicly released F5-TTS base model~\citep{f5tts} with its default inference recipe: $32$ ODE solver steps, CFG~\citep{cfg} scale $2.0$, and sway sampling.
We generate $10$ BoN candidates per utterance; for each $N\in\{3,5,10\}$ we take the leading $N$ candidates and select the one with the lowest joint WER+CER score against the reference transcript.

\textbf{Verifiers.}
We compare three single-model verifiers and one three-way ensemble.
The single verifiers are a wav2vec~2.0 base model (95M parameters; \emph{w2v2-base}), a small Distil-Whisper (166M; \emph{distil-sm}), and a large Distil-Whisper (756M; \emph{distil-v3}).
\emph{ens3} denotes the three-way rank-average of these three verifiers.
In the scaling study (\S\ref{sec:scaling}) we additionally consider two cross-family rank ensembles that aggregate only the cross-family pair \{w2v2-base, distil-v3\}.

\textbf{Evaluators.}
For the main results (\S\ref{sec:main}) we score every candidate with four ASR evaluators spanning three families: a Whisper-large-v3 evaluator served via the faster-whisper runtime (\emph{fwhisper-lgv3}, the official F5-TTS evaluator~\citep{f5tts}); a large wav2vec~2.0 model (\emph{w2v2-lv60}); the same Distil-Whisper used as a verifier (\emph{distil-sm}); and a large HuBERT model (\emph{hubert-lg}).
For the CKA--WER analysis (\S\ref{sec:analysis}) we additionally include Whisper-medium (\emph{whisper-med}) as a fourth evaluator to expand the pool of evaluator pairs.

\textbf{Data \& normalization.}
We evaluate on the LibriSpeech-PC test-clean cross-sentence subset~\citep{librispeechpc} ($1127$ samples, $4$--$10$\,s reference) and run the F5-TTS official evaluation pipeline verbatim, applying its default text normalization uniformly to all four ASR outputs. Whisper's English-specific normalizer is not applied separately, to keep the cross-ASR WER comparison on the same footing.

\textbf{Quality.}
Speaker similarity SIM-o is the cosine of WavLM~\citep{wavlm} speaker embeddings; reference and synthesized audio are resampled symmetrically from $24$\,kHz to $16$\,kHz before scoring.
Naturalness is the UTMOS~\citep{utmos} score from the standard $22$-strong checkpoint.

\textbf{Reproducibility.}
All models are used at their default public revisions; code, decoding settings, and evaluation scripts are available at \url{https://github.com/yu1012/BoN-TTS}.

\section{Main Results}
\label{sec:main}

\subsection{Results under the F5-TTS Evaluator}

\begin{table}[tb]
\centering
\caption{LibriSpeech-PC test-clean under the official fwhisper-lgv3 evaluator. ``rel'' is WER reduction vs.\ F5-TTS; ``$p$'' is a two-sided paired permutation test against F5-TTS (10{,}000 permutations).}
\label{tab:headline}
\small
\setlength{\tabcolsep}{3.5pt}
\begin{tabular}{lccccc}
\toprule
Method & WER & CER & RTF & rel & $p$ \\
\midrule
F5-TTS (baseline) & 2.06 & 0.71 & 0.190 & --- & --- \\
BoN w2v2-base & 1.99 & 0.62 & 0.383 & $-3.5\%$ & 0.31 \\
BoN distil-sm & 2.04 & 0.59 & 0.388 & $-1.0\%$ & 0.93 \\
\textbf{BoN distil-v3} & \textbf{1.88} & \textbf{0.53} & 0.379 & $\mathbf{-8.7\%}$ & \textbf{0.030} \\
BoN ens3      & 1.91 & 0.54 & 0.429 & $-7.2\%$ & 0.057 \\
\bottomrule
\end{tabular}
\end{table}

Our single-shot F5-TTS baseline reaches $2.06\%$ WER, within $0.36$\,pp of the $2.42\%$ multi-seed mean reported in the F5-TTS paper~\citep{f5tts}---a single-shot vs.\ multi-seed-mean gap, not a pipeline discrepancy.
BoN with distil-v3 drives WER to $1.88\%$ ($-8.7\%$, $p{=}0.030$) at RTF $0.379$, the only significant single-verifier reduction; ens3 is a close runner-up ($-7.2\%$, $p{=}0.057$). The w2v2-base verifier improves the point estimate ($-3.5\%$) but is not significant ($p{=}0.31$), and distil-sm is flat ($p{=}0.93$).
Speaker similarity (SIM-o) stays within $\pm 0.0006$ and UTMOS within $\pm 0.005$ of F5-TTS across all configurations (\S\ref{sec:quality}), so the WER gain comes at no measurable quality cost.

\subsection{Cross-Evaluator Triangulation}

\begin{table}[tb]
\centering
\caption{Cross-evaluator WER (\%) on LibriSpeech-PC test-clean. The same generated audio is evaluated by ASR checkpoints from different families (aliases defined in \S\ref{sec:setup}); the preferred verifier (bold) changes with the evaluator.}
\label{tab:xeval}
\small
\setlength{\tabcolsep}{4pt}
\begin{tabular}{lccc}
\toprule
Method & fwhisper-lgv3 & w2v2-lv60 & hubert-lg \\
\midrule
F5-TTS (baseline)      & 2.06 & 1.52 & 1.92 \\
BoN w2v2-base          & 1.99 & \textbf{1.41} & \textbf{1.74} \\
BoN distil-sm          & 2.04 & 1.50 & 1.77 \\
BoN distil-v3          & \textbf{1.88} & 1.45 & \textbf{1.74} \\
BoN ens3               & 1.91 & 1.45 & 1.76 \\
\bottomrule
\end{tabular}
\end{table}

\textbf{Same-family preference.} Under fwhisper-lgv3 (Table~\ref{tab:xeval}), the Distil-Whisper verifier (distil-v3) wins; under w2v2-lv60, the wav2vec~2.0 verifier (w2v2-base) wins; under hubert-lg, w2v2-base again ranks at the top alongside distil-v3.

\textbf{Magnitude of the swing.} The distil-v3 BoN reduction is $0.18$\,pp under fwhisper-lgv3 but only $0.07$\,pp under w2v2-lv60, reversing the verifier ranking between Whisper and wav2vec~2.0 evaluators. A reader given only the F5-TTS official evaluator would prefer distil-v3 BoN; one given only w2v2-lv60 would prefer w2v2-base BoN. Both are correct conditional on their evaluator, and incompatible.

\subsection{Oracle Headroom and Recovery}

\begin{table}[tb]
\centering
\caption{Oracle headroom on LibriSpeech-PC test-clean with $N{=}3$ candidates per sample. \emph{Single-shot} $=$ first candidate; \emph{Oracle} $=$ per-sample best of $3$. The bottom block reports recovery $(\mathrm{Single}{-}\mathrm{BoN})/(\mathrm{Single}{-}\mathrm{Oracle})$ in \%; same-family BoN recovers $2$--$3{\times}$ more than cross-family BoN on the same audio. The $0.02$\,pp gap between Single-shot $=2.04$ here and F5-TTS $=2.06$ in Tables~\ref{tab:headline}/\ref{tab:scaling} is Whisper batching non-determinism (decoded in an $N{=}3$ batch vs.\ a single utterance); both rows describe identical synthesized audio.}
\label{tab:oracle}
\small
\setlength{\tabcolsep}{4pt}
\begin{tabular}{lccc}
\toprule
                       & fwhisper-lgv3 & w2v2-lv60 & hubert-lg \\
\midrule
Single-shot            & 2.04 & 1.52 & 1.94 \\
\textbf{Oracle ($N{=}3$)} & \textbf{1.42} & \textbf{1.09} & \textbf{1.18} \\
Headroom (pp)          & 0.63 & 0.43 & 0.76 \\
\midrule
\multicolumn{4}{l}{\emph{Recovery (\%) of oracle headroom:}} \\
BoN w2v2-base          & 7.9 & \textbf{26.1} & 27.1 \\
BoN distil-sm          & 0.0 & 6.8 & 22.6 \\
\textbf{BoN distil-v3} & \textbf{26.0} & 18.2 & 27.1 \\
BoN ens3               & 20.5 & 17.0 & 24.5 \\
\bottomrule
\end{tabular}
\end{table}

To bound how far BoN can go at $N{=}3$, we synthesize all three candidates per sample and transcribe each with every evaluator (Table~\ref{tab:oracle}, top block).
On the official evaluator, an oracle verifier would drive WER to $1.42\%$, below F5-TTS's reported $\sim$$1.5$--$1.7\%$~\citep{f5tts}: headroom for verifier improvement is substantial.
The bottom block makes the $2$--$3{\times}$ family swing precise: under fwhisper-lgv3, distil-v3 recovers \textbf{26.0\%} of headroom vs.\ only $7.9\%$ for cross-family w2v2-base ($3.3\times$); under w2v2-lv60, w2v2-base recovers \textbf{26.1\%} vs.\ $18.2\%$ for cross-family distil-v3 ($1.4\times$). Recovery is a property of the (verifier, evaluator) pair, not the verifier alone.

\subsection{Scaling and Bias-Corrected Ensembles}
\label{sec:scaling}

To test whether the confound persists at higher $N$ and whether cross-family aggregation mitigates it, we synthesize $N{=}10$ candidates per sample and add six BoN configurations: single wav2vec~2.0 and Distil-Whisper verifiers at $N\in\{5,10\}$, plus two cross-family rank ensembles over the (w2v2-base, distil-v3) pair.

\begin{itemize}
\setlength{\itemsep}{2pt}\setlength{\parskip}{0pt}\setlength{\topsep}{2pt}
\item \textbf{rank-avg.} Each verifier independently ranks the $N$ candidates by WER+CER score. We then pick the candidate with the lowest average rank across the two families.
\item \textbf{max-rank.} We pick the candidate with the lowest worst-case rank, requiring it to rank well in both families. This conjunctive form explicitly penalizes single-family inflation.
\end{itemize}

\begin{figure}[t]
\centering
\begin{tikzpicture}
\begin{axis}[
  width=0.92\linewidth, height=3.7cm,
  xlabel={\small $N$ (BoN candidates)},
  ylabel={\small Mean WER (\%)},
  xtick={3,5,10},
  ymin=1.55, ymax=1.85,
  legend style={font=\scriptsize, at={(0.5,-0.34)}, anchor=north,
                legend columns=3, draw=none, fill=none,
                /tikz/every even column/.append style={column sep=10pt}},
  tick label style={font=\scriptsize},
  label style={font=\scriptsize},
  grid=major, grid style={gray!20},
]
\addplot[mark=square*, color=red!70!black, thick] coordinates
  {(3,1.71)(5,1.79)(10,1.70)}; \addlegendentry{w2v2-base}
\addplot[mark=*, color=blue!70!black, thick] coordinates
  {(3,1.69)(5,1.68)(10,1.61)}; \addlegendentry{distil-v3}
\addplot[mark=triangle*, color=green!50!black, thick] coordinates
  {(3,1.73)(5,1.65)(10,1.61)}; \addlegendentry{rank-avg}
\addplot[mark=diamond*, color=orange!80!black, thick] coordinates
  {(3,1.71)(5,1.63)(10,1.61)}; \addlegendentry{max-rank}
\addplot[dashed, color=gray] coordinates {(3,1.83)(10,1.83)};
\addlegendentry{F5-TTS}
\end{axis}
\end{tikzpicture}
\vspace{1ex}
\caption{Mean WER (averaged over fwhisper-lgv3, w2v2-lv60, hubert-lg) vs.\ $N$, from Table~\ref{tab:scaling}. Both cross-family rank ensembles (rank-avg, max-rank) at $N{=}5$ reduce WER under all three evaluators simultaneously, while the single w2v2-base verifier regresses on fwhisper-lgv3 at $N{=}5$. Cross-family ensembles scale monotonically with $N$.}
\label{fig:scaling}
\end{figure}
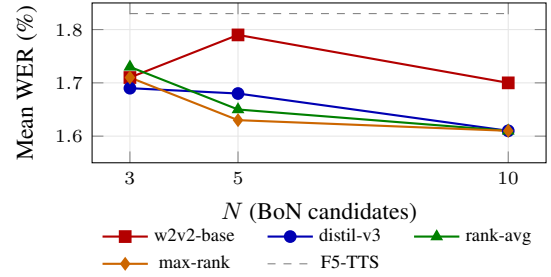

\begin{table}[tb]
\centering
\caption{Scaling $N$ on LibriSpeech-PC test-clean. Per-evaluator WER (\%) under three evaluators (aliases as defined in \S\ref{sec:setup}). The $p$-value column reports a two-sided paired permutation test against F5-TTS under \emph{fwhisper-lgv3} (10{,}000 permutations). ``$\uparrow$'' marks a regression vs.\ the same row at the preceding $N$.}
\label{tab:scaling}
\small
\setlength{\tabcolsep}{2.6pt}
\resizebox{\columnwidth}{!}{%
\begin{tabular}{lccccc}
\toprule
Method ($N$) & fwhisper-lgv3 & w2v2-lv60 & hubert-lg & mean & $p$ \\
\midrule
F5-TTS (baseline) & 2.06 & 1.52 & 1.92 & 1.83 & --- \\
\midrule
w2v2-base ($3$)    & 1.99          & 1.41          & 1.74          & 1.71 & 0.05 \\
w2v2-base ($5$)    & 2.20$\uparrow$& 1.41          & 1.75          & 1.79 & 0.97 \\
w2v2-base ($10$)   & 1.98          & 1.40          & 1.72          & 1.70 & 0.08 \\
\midrule
distil-v3 ($3$)  & 1.88          & 1.45          & 1.74          & 1.69 & \textbf{$<\!.001$} \\
distil-v3 ($5$)  & 1.80          & 1.48          & 1.75          & 1.68 & \textbf{$<\!.001$} \\
distil-v3 ($10$) & \textbf{1.72} & 1.44          & 1.68          & \textbf{1.61} & \textbf{$<\!.001$} \\
\midrule
rank-avg ($3$)   & 2.01          & 1.43          & 1.74          & 1.73 & 0.17 \\
rank-avg ($5$)   & 1.90          & \textbf{1.40} & 1.66          & 1.65 & \textbf{.001} \\
rank-avg ($10$)  & 1.81          & 1.41          & \textbf{1.60} & 1.61 & \textbf{$<\!.001$} \\
\midrule
max-rank ($3$)   & 1.99          & 1.42          & 1.73          & 1.71 & 0.12 \\
max-rank ($5$)   & 1.80          & 1.43          & 1.67          & 1.63 & \textbf{$<\!.001$} \\
max-rank ($10$)  & 1.80          & \textbf{1.40} & 1.62          & 1.61 & \textbf{$<\!.001$} \\
\bottomrule
\end{tabular}%
}
\end{table}

\textbf{Findings} (Figure~\ref{fig:scaling}).
(i) The single cross-family verifier (w2v2-base) under fwhisper-lgv3 regresses from $1.99\%$ at $N{=}3$ to $2.20\%$ at $N{=}5$ ($p{=}0.97$): more candidates expose a same-family-evaluator penalty.
(ii) The same-family verifier (distil-v3) scales monotonically on its own family ($1.88\%\rightarrow 1.72\%$ for $N{=}3{\to}10$, $p{<}.001$).
(iii) Cross-family rank ensembles match the single best on mean WER and are the most cross-evaluator-robust. The rank-avg configuration at $N{=}5$ is the only one reaching $p{<}0.05$ under all three evaluators at once ($.001/.022/.0002$); at $N{=}10$, rank-avg ties distil-v3 at mean $1.61\%$ while winning the hubert-lg column ($1.60\%$).

\textbf{Practical rule.} With a known same-family evaluator, use distil-v3 at $N{=}10$. Otherwise, use rank-avg or max-rank over the cross-family pair at $N{=}5$ for the best compute--robustness frontier.

\section{Analysis: Toward a Mechanism for the Confound}
\label{sec:analysis}

\begin{table}[tb]
\centering
\caption{Linear CKA (mean-pooled last hidden state) between audio encoders of six ASR checkpoints on $500$ F5-TTS-synthesized waveforms (a random subset of LibriSpeech test-clean). Aliases as in \S\ref{sec:setup}: wav2vec~2.0 family (\emph{w2v2-base}, \emph{w2v2-lv60}), Whisper family (\emph{distil-sm}, \emph{distil-v3}, \emph{whisper-med}; the latter three are Distil-Whisper or Whisper checkpoints), HuBERT family (\emph{hubert-lg}).}
\label{tab:cka}
\small
\setlength{\tabcolsep}{3pt}
\resizebox{\columnwidth}{!}{%
\begin{tabular}{lcccccc}
\toprule
& \rotatebox{60}{w2v2-base} & \rotatebox{60}{w2v2-lv60} & \rotatebox{60}{distil-sm} & \rotatebox{60}{distil-v3} & \rotatebox{60}{whisper-med} & \rotatebox{60}{hubert-lg} \\
\midrule
w2v2-base    & 1.00 & 0.65 & 0.14 & 0.14 & 0.15 & 0.67 \\
w2v2-lv60 & 0.65 & 1.00 & 0.06 & 0.05 & 0.06 & 0.55 \\
distil-sm    & 0.14 & 0.06 & 1.00 & \textbf{0.98} & \textbf{0.98} & 0.39 \\
distil-v3    & 0.14 & 0.05 & \textbf{0.98} & 1.00 & \textbf{0.98} & 0.40 \\
whisper-med  & 0.15 & 0.06 & \textbf{0.98} & \textbf{0.98} & 1.00 & 0.40 \\
hubert-lg    & 0.67 & 0.55 & 0.39 & 0.40 & 0.40 & 1.00 \\
\bottomrule
\end{tabular}%
}
\end{table}

To test whether the confound reduces to representation similarity, we measure linear CKA~\citep{cka} between audio encoders of six ASR checkpoints on $500$ F5-TTS-synthesized waveforms (a random subset of LibriSpeech test-clean). Table~\ref{tab:cka} confirms three intuitive clusters: Whisper--Distil-Whisper (CKA $0.97$--$0.98$ internally), wav2vec~2.0 / HuBERT ($0.55$--$0.67$), and near-zero cross-cluster.

\begin{figure}[tb]
\centering
\includegraphics[width=\linewidth]{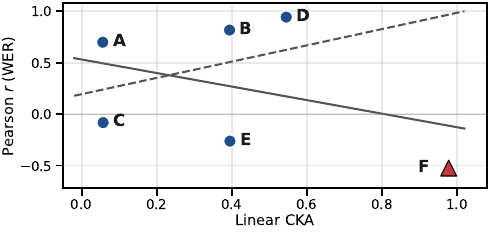}
\caption{CKA does not predict WER agreement across the $\binom{4}{2}{=}6$ evaluator pairs from $\{$distil-sm, w2v2-lv60, whisper-med, hubert-lg$\}$. Pearson $r$ is computed over $13$ BoN configurations on a $500$-sample pilot.
Pairs: \textbf{A}, distil-sm $\leftrightarrow$ w2v2-lv60; \textbf{B}, distil-sm $\leftrightarrow$ hubert-lg; \textbf{C}, w2v2-lv60 $\leftrightarrow$ whisper-med; \textbf{D}, w2v2-lv60 $\leftrightarrow$ hubert-lg; \textbf{E}, whisper-med $\leftrightarrow$ hubert-lg; \textbf{F}, distil-sm $\leftrightarrow$ whisper-med (same-family, red triangle).
The all-6 trend ($r{=}{-}0.36$, solid) is driven by the same-family outlier F ($r_{\mathrm{WER}}{=}{-}0.52$ at CKA $0.978$). Excluding F, the trend reverses to $r{=}{+}0.32$ (dashed), confirming that CKA does not monotonically predict WER agreement.}
\label{fig:cka-disagree}
\end{figure}

The naive prediction (high CKA implies similar WER rankings) is not supported (Figure~\ref{fig:cka-disagree}). The Whisper-family pair (distil-sm $\leftrightarrow$ whisper-med, CKA $0.978$) is anti-correlated ($r{=}{-}0.52$), while the wav2vec~2.0/HuBERT pair (CKA $0.55$) agrees nearly perfectly ($r{=}{+}0.94$). Pearson(CKA, $r$) is $-0.36$ over the six pairs but reverses to $+0.32$ once the same-family outlier is removed: the trend is single-point-driven.

The pattern is consistent not with representation similarity but with an \textbf{identity- or lineage-level coupling effect}: a verifier's selections are disproportionately favored by evaluators sharing its ASR lineage, while an independent same-family member returns an uninflated score---structurally analogous to LLM-as-a-judge self-bias~\citep{llmasjudge}. Disentangling identity- from family-level coupling is left to future work.

\subsection{Quality and cross-evaluator robustness}
\label{sec:quality}

SIM-o and UTMOS at $N{=}3$ stay within $\pm 0.001$ and $\pm 0.005$ of F5-TTS ($0.9426$/$3.879$), with no measurable degradation under either predictor. Both may be near their resolution limit (UTMOS saturates around $4.0$), so human-MOS / NISQA / UTMOSv2 triangulation~\citep{mittag2021nisqa,utmosv2} is deferred. For cross-evaluator robustness, rank-avg and max-rank at $N{=}5$ reduce WER under all three evaluators simultaneously, while distil-v3 at $N{=}10$ stays flat on w2v2-lv60. The (w2v2-base, distil-v3) pair was fixed in advance as the largest publicly available verifier in each family, prior to any cross-evaluator analysis.

\section{Discussion and Conclusion}
\label{sec:discussion}

Any BoN comparison reported under a single ASR evaluator is potentially confounded; we recommend reporting WER under at least two ASR families with disjoint training lineages. Our \textbf{cross-family rank ensembles} realize this principle on the selection side: they yield the most cross-evaluator-robust BoN configurations we test, reaching $1.61\%$ mean WER at $N{=}10$ ($-12\%$ vs.\ F5-TTS); the largest reduction under the official F5-TTS evaluator alone, $2.06\%\rightarrow1.72\%$ ($-16.5\%$), comes from the best same-family verifier. The $N{=}3$ oracle WER of $1.42\%$ leaves substantial headroom for verifier-design work (e.g., adversarial reweighting against in-family inflation).

\textbf{Limitations.} Our study uses one TTS backbone (F5-TTS) on LibriSpeech-PC test-clean; generalization to other backbones (CosyVoice~2, MaskGCT) and a fourth ASR family (Parakeet, Canary), plus human-MOS triangulation, is left to future work.

\bibliography{refs}
\bibliographystyle{icml2026}

\end{document}